# Stress Detection Using PPG Signal and Combined Deep CNN-MLP Network

Yasin Hasanpoor
*Advanced Service Robots (ASR) Laboratory*
*Department of Mechatronics Engineering*
*Faculty of New Sciences and Technologies*
*University of Tehran*
Tehran, Tehran, Iran
yasin.hasanpoor@ut.ac.ir

Koorosh Motaman
*Advanced Service Robots (ASR) Laboratory*
*Department of Mechatronics Engineering*
*Faculty of New Sciences and Technologies*
*University of Tehran*
Tehran, Tehran, Iran
koorosh.motaman@ut.ac.ir

Bahram Tarvirdizadeh
*Advanced Service Robots (ASR) Laboratory*
*Department of Mechatronics Engineering*
*Faculty of New Sciences and Technologies*
*University of Tehran*
Tehran, Tehran, Iran
bahram@ut.ac.ir

Khalil Alipour
*Advanced Service Robots (ASR) Laboratory*
*Department of Mechatronics Engineering*
*Faculty of New Sciences and Technologies*
*University of Tehran*
Tehran, Tehran, Iran
k.alipour@ut.ac.ir

Mohammad Ghamari
*Department of Electrical and Computer Engineering*
*Kettering University*
Flint, Michigan, USA
mghamari@kettering.edu

*Abstract*— Stress has become a fact in people's lives. It has a significant effect on the function of body systems and many key systems of the body including respiratory, cardiovascular, and even reproductive systems are impacted by stress. It can be very helpful to detect stress episodes in early steps of its appearance to avoid damages it can cause to body systems. Using physiological signals can be useful for stress detection as they reflect very important information about the human body. PPG signal due to its advantages is one of the mostly used signal in this field. In this research work, we take advantage of PPG signals to detect stress events. The PPG signals used in this work are collected from one of the newest publicly available datasets named as UBFC-Phys and a model is developed by using CNN-MLP deep learning algorithm. The results obtained from the proposed model indicate that stress can be detected with an accuracy of approximately 82 percent.

## I. Introduction

Stress has become an inseparable part of modern life. Many people experience stress on daily basis due to a variety of reasons. Each person may experience stress at different intensities and for different reasons [1]. Stress can be acute or chronic, both leading to a variety of undesirable psychological, physical, and behavioral problems [2], [3], [4]. According to research studies, early detection of acute stress can assist in mitigating chronic stress [5], [6]. Therefore, it is vital to detect stress early to prevent future major problems. Until today, many attempts have been made for detecting acute stress such as in [7]. Authors in [8] showed that it is possible to detect acute stress. For instance, Authors in [8] attempted to find physiological or behavioral markers for stress by analyzing correlations between physiological or behavioral data to find specific features associated with stress. To achieve this goal, they used machine learning (ML) methods for stress classification. Stress can be detected in real-time as shown in [9]. Authors in [9] Fabricated a real-time stress detection portable device equipped with PPG and GSR sensors and then extracted and classified signal features with ML algorithms in order to detect three different levels of stress.

One of the most recognized and practical approach for real-time stress detection is using human body's physiological signals [10]. Stress has shown to have significant influence on human body's physiological systems [11]. In addition, according to [12], physiological signals are capable of indicating the state of physiological systems very well. Therefore, analyzing physiological signals of human body can help in detection of acute stress [5]. A variety of physiological signals can be received from body including electrocardiogram (ECG), Electromyography (EMG), Heart Rate (HR), Photoplethysmography (PPG), galvanic skin response (GSR) and Blood Pressure (BP). It must be noted that, each of these signals require a specific sensor and or a particular method for reception of signal. Among all these physiological signals, PPG signal is more recognized because of its advantages in healthcare [13], [14]. PPG signal can be collected in a cost-effective non-invasive manner [9]. Moreover, in contrast to other signals, it contains valuable health-related information [15]. PPG sensor can be rigid [16], [17] or printed flexible [18] and can be incorporated into wearable devices [19], [20].

Until now, many studies have been done in the field of stress detection and emotion recognition using physiological signals [21]. These studies have some major differences as mentioned in the following. The greatest difference in these studies consist of the type of signals in addition to the model of the algorithms they employ. For example, authors in [22] used EEG signals, in [23] used EDA signals, in [24] used HRV signals, in [25] used blood pressure signals, in [26] used HR signals in addition to many other works that used other types of signals. This study uses PPG signals as the basis for stress detection. Similar studies in this field have been done in the past as shown in [27].

As algorithm selection is another important aspect of this work, specific algorithms have been used for creating the model. Different algorithms such as logistic regression [28], Support Vector Machine (SVM) [29], k-nearest neighbors (KNN) [30], decision tree [31], Multilayer perceptron (MLP) [32] and many other more algorithms have been used in the previous studies. There are many other algorithms and examples that can be found in [10], [21]. However, most of algorithms used for this type of work have been based on machine learning techniques. The use of deep learning algorithms has been neglected and rarely used. A few studies such as work that has been done in [33] used deep learning algorithms. In contrast to the machine learning algorithms that uses extracted features of signal for training the machine, deep learning algorithms can use the signal itself for training the machine. Therefore, in this work, deep learning algorithms have been used for stress detection. CNN-MLP algorithm [34] is used to develop a stress detection model in this work.

Another important parameter for developing a stress detection model is the quality of the dataset. Dataset can affect the applicability of the model. Although not many high-quality datasets have been available in this field, in order to perform this work, one of the most widely used and more accessible dataset were used. In this work, UBFC-Phys [35] dataset was used. UBFC-Phys is used because of being more standard in comparison to other datasets. Moreover, it consists of a higher number of participants.

## II. THE DATASET

As mentioned before dataset can affect the applicability of the model for use in daily life. As a general rule, the larger the dataset and the greater number of participants, the more general the model will be. Therefore, the UBFC-Phys dataset [35] has very good specifications from this angle of view. For this work UBFC-Phys is used for training the deep learning model.

### A. UBFC-PHYS

The UBFC-Phys dataset is a multimodal physiological and visual dataset. The data in this dataset is collected in different situations. The data in UBFC-Phys consists of two main parts: The first part consists of the recordings of subject's facial video and the second part of the data consists of physiological signals obtained from wrist-worn device "Empatica E4". The blood volume pulse (BVP) signal (also known as PPG) and the electrodermal activity (EDA) signal (also known as GSR) are the signals that was recorded using "Empatica E4" wrist sensor.

The Dataset was recorded in an experiment that contains three tasks for each participant: a rest task (T1) that gets the subjects baseline, a speech task (T2) and an arithmetic task (T3). The last two tasks can be stressful tasks.

3 minutes of each task has recorded for each subject and total of 9 minutes of signal is available for each of individual. There are a total of 56 participants in this test. This work only uses the rest and the speech tasks containing PPG signals for training the deep learning model. T1 and T2 were defined as stress and non-stress classes respectively. Fig. 1 is a 10s section of the 9th person's PPG to see how stress can impact PPG signal. It's needed to note that the PPG data was collected with sample rate of 64 Hz.

## III. PREPARING DATASET

As mentioned before, only PPG signals were extracted from the UBFC-Phys dataset. For this work, we used rest task data for class 1 (as non-stress) and used speech task data for class 2 (as stress) and did not consider the arithmetic task data. As data were collected at 64 Hz sample rate and every task of every participant lasted 3 minutes, each task data is a 11520 long vector. We concatenated all the vectors of each class data as you can see in Fig. 2.

| Class 1 | Subject1 [1x11520] | Subject2 [1x11520] | ......... | Subject56 [1x11520] |
| Class 2 | Subject1 [1x11520] | Subject2 [1x11520] | ......... | Subject56 [1x11520] |

Figure 2. Arrangement of PPG signals as the CNN input

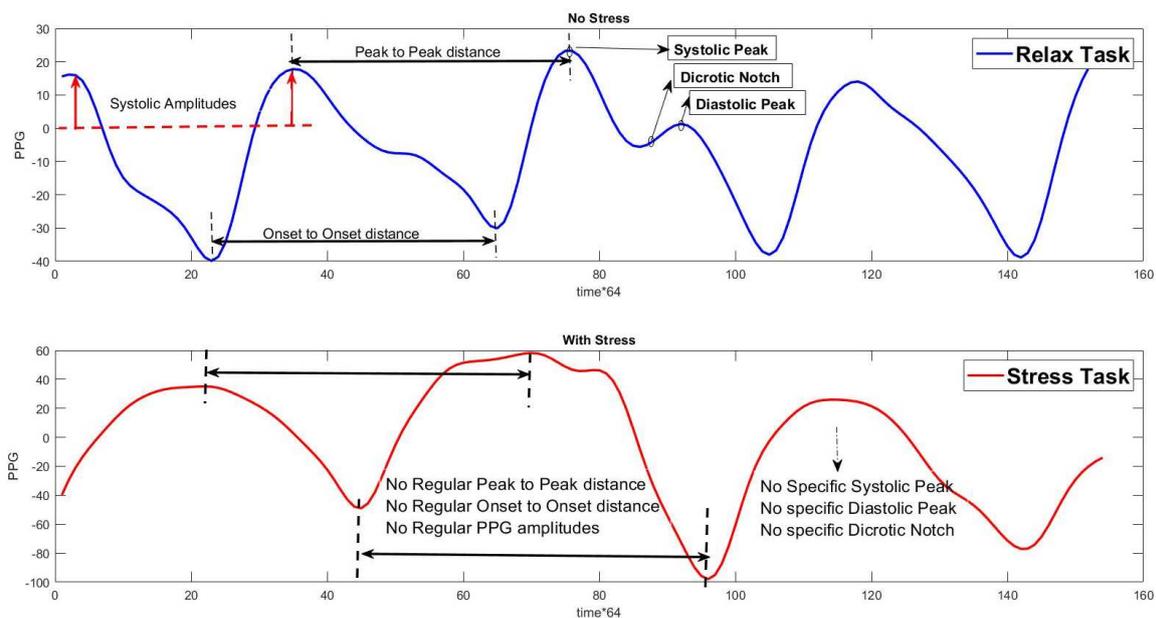

Figure 1. impact of stress on PPG signals; Normal PPG has a rhythmic structure with equal peak to peak times and distinguishable Systolic and diastolic points, while stress disrupt these properties and change amplitudes, peak to peak times and other specifications

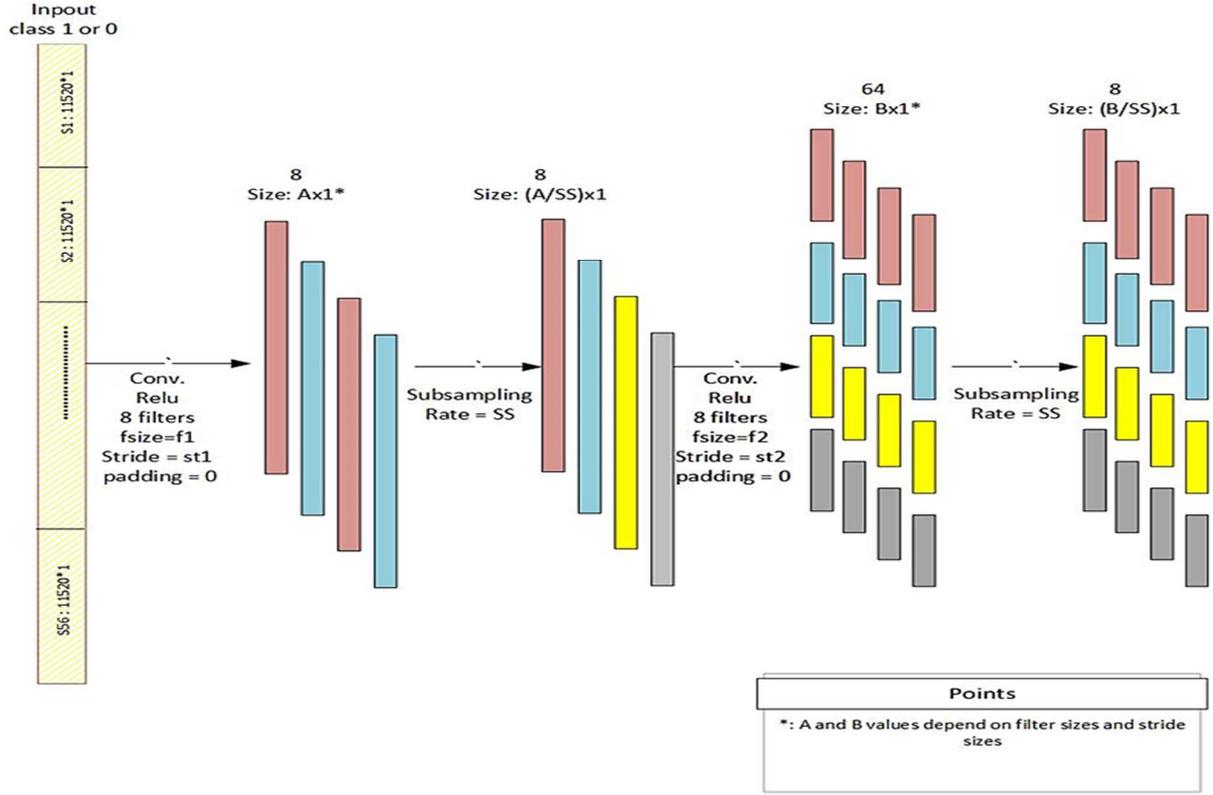

Figure 3. Data flow through 2 CNN layers

After that, we normalized the data between +1 and -1. Even though signals contain unwanted noises such as motion artifacts (M.A.), sensor's noises and other unavoidable noises. No further preprocessing was implemented. While using "ChebyshevII" filter, which was implemented at first, caused 2% reduction in accuracy on average. this shows that the CNN can detect noises by itself better. Thereby, without any further preprocessing and filtering, class vectors were directly fed into the neural networks.

## IV. CONVOLUTIONAL NEURAL NETWORK DESIGN AND IMPLEMENTATION

It is obvious that our input data is in form of vectors while the CNN network (that is a part of our network) is for image classification in the original version and designed to get two-dimensional data as input [34]. we need a structure that can get vectors as input. The design of the CNN we used in this paper was inspired by the structure used in [34] With differences which are back-propagation training in CNN layers beside fully connected ones (I.e. MLP) and one-dimensional vector input. In the following, we shall discuss how data process through CNN layers and then how MLP gets the CNN output for classification.

Fig. 3 shows briefly how the data flow in 2 CNN layers. This process repeats in CNN layers until the last layer of CNN and before the first MLP layer. During the forward propagation, the input vector of each layer will be obtained by the sum of the final outputs of the previous layer neurons convolved with the related kernels as equation 1, where x is the output of the neuron, b is the bias, w is the weight convolved with the output of the previous neuron.

$$x_k^l = b_k^l + \sum_{i=1}^{N_{l-1}} conv1D(w_{ik}^{l-1} \cdot s_i^{l-1}) \quad (1)$$

By arranging input in 1D vectors, Fig. 4 shows briefly how the frame changes and is convolved with filters. shorter strides make the model more accurate as was concluded in the results section. The results of convolution operations are summed with the bias and make one neuron in the next CNN layer. Moving the frame makes new operations and thereby new neurons. We can change the stride value and size of frames as many as we want. These are the parameters of the structure and the optimal values are shown in the Result section.

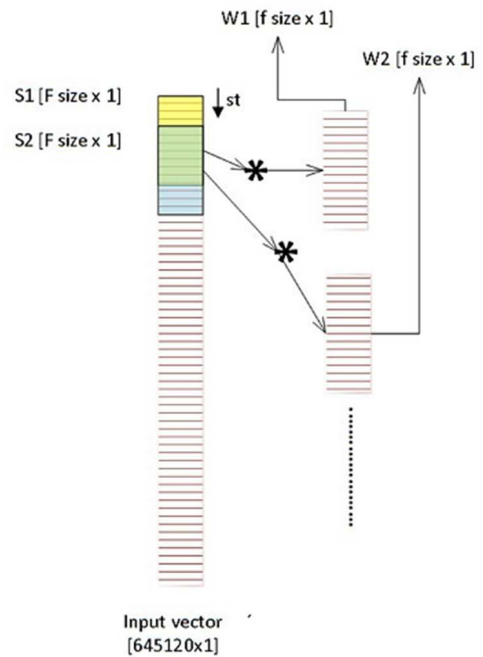

Figure 4. frame striding and convolution of weights and signals

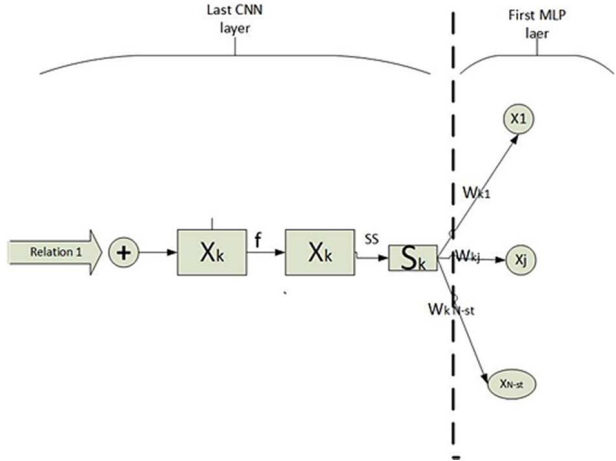

Figure 5. CNN output to 1st MLP layer

The last layer of CNN is connected to the first layer of MLP. Therefore, the CNN output should be scaler. Our design is adaptive and the output layer of CNN is obtained by the sum of convolution results. In addition, the subsampling rate with filter sizes is set to the input vector. Fig. 5 shows how CNN delivers the output to MLP. As detailed in this figure, the sum of convolved signals with their kernels (Eqn. 1) makes the input signal of the CNN. After the activation function, the subsampled signal ($S_k$) is created as the output signal of CNN. Finally, $S_k$ signals convolve with the kernels and make the input signals of MLP. All these processes repeat for other neurons. Thus far, the Design of the network was described. We shall discuss the structure parameters and a brief summary of implementation on MATLAB.

numbers of filters are decided to be 8 and 5 in all CNN and MLP layers respectively since more filters cause no accuracy improvement, but also longer time of runs. Changes in other parameters mentioned in Table I, cause changes in the accuracy and performance of the network as summarized in Table II.

TABLE I. STRUCTURE PARAMETER SYMBOLS AND DOMAIN OF CHANGES

| Parameter name | symbol | Tested values |
|---|---|---|
| number of CNN layers | n | 3,2 |
| number of MLP layers | m | 3,2 |
| frame sizes | Fsize | 2048,1024,512,256,128,64 |
| filter sizes | fsize | 128, 64, 32 |
| subsampling rates | SS | 8,6,4,2 |
| stride | st | 24,12,10,8,5,2 |

Table I shows values that were tested for each parameter. While other parameters were kept unchanged. Implementation of the network described in the following.

After Data normalization, to be ensured that the model will not be "stuck" with too many bad batches, shuffling of data after each epoch was considered. Data was divided in to parts of train and test with 40% for train. Initial values of network mentioned in Table I were set to be 3, 3, 128, 32, 2, 24 respectively from above. Stop conditions are in the following; the number of epochs up to 200 or least mean square of delta (in BP) reaches 0.001 or best fit occurs (prevent underfitting or overfitting). The model was run on MATLAB and runtime was 10mins on average by means of CPU intel core i7-7500U.

## V. RESULTS

A number of different structures were designed and tested to get the best result possible from training the CNN-MLP network using the UBFC-Phys dataset's PPG Signals for stress Classification. Parameters such as the number of layers in each CNN and MLP part, frame size, filter size, subsampling rate, and stride were optimized by trial and error. In the best result, we could achieve 82% accuracy. The results

TABLE II. CHANGES OF STRUCTURE PARAMETERS AND RESULTS OF ACCURACIES ON TRAINING AND TESTING

| No. of classes | No. of CNN layers (n) | No. of MLP layers (m) | Frame size (F size) | Filter size (f size) | Subsampling rate (SS) | stride | Training accuracy | Testing accuracy |
|---|---|---|---|---|---|---|---|---|
| 2 | 3 | 2 | 64 | 16 | 2 | 24 | 77% | 75% |
| 2 | 3 | 2 | 128 | 32 | 2 | 24 | 80% | 77% |
| 2 | 2 | 2 | 512 | 64 | 2 | 24 | 78% | 78% |
| 2 | 2 | 2 | 1024 | 128 | 2 | 24 | 86% | 72% |
| 2 | 2 | 2 | 1024 | 512 | 4 | 24 | 85% | 73% |
| 2 | 2 | 2 | 1024 | 512 | 4 | 12 | 89% | 80% |
| 2 | 2 | 2 | 1024 | 512 | 4 | 5 | 85% | 75% |
| 2 | 2 | 2 | 1024 | 512 | 6 | 5 | 88% | 81% |
| 2 | 2 | 2 | 1024 | 512 | 8 | 2 | 92% | 82% |

of different settings and structures of the network can be found in Table II.

## VI. CONCLUSIONS

In this work we developed a combined CNN-MLP deep network for UBFC-Phys dataset. The UBFC-Phys dataset consisted of raw ppg signals. Different goals were targeted in this work. First, the applicability of a new dataset for stress detection was evaluated in this work. Second, the ability of the deep learning models, specifically CNN-MLP, was tested for developing stress detection models based on PPG signal. In this work, an accuracy of 82% was achieved. By considering this fact that only raw PPG signals was used for this work, and authors believe that this accuracy pretty good. PPG signals were only normalized. experimental results show that the CNN can detect a part of motion artifact noises by itself. Table II describe briefly how parameters were optimized for the proposed model. As can be seen, increasing frame sizes and filter sizes in parallel caused to obtain better results using the proposed model. In addition, higher subsampling rates made the proposed model be improved and perform faster in runtime. Finally, making shorter strides helped the proposed model to reach better accuracies in both the training and testing of classifiers.